\documentclass[10pt,twocolumn,letterpaper]{article}

\usepackage{wacv}
\usepackage{times}
\usepackage{epsfig}
\usepackage{graphicx}
\usepackage{amsmath}
\usepackage{amssymb}
\usepackage{booktabs}
\usepackage[accsupp]{axessibility} 

\usepackage[ruled,vlined, linesnumbered]{algorithm2e}
\usepackage{multirow}

\usepackage{booktabs}
\usepackage{floatrow}
\usepackage{wrapfig}
\usepackage{caption}
\usepackage{float}
\newfloatcommand{capbtabbox}{table}[][\FBwidth]
\usepackage{blindtext}

\newcommand{\norm}[1]{\left\lVert#1\right\rVert}

\def\wacvPaperID{657} 

\wacvalgorithmstrack   

\wacvfinalcopy 

\ifwacvfinal
\usepackage[breaklinks=true,bookmarks=false]{hyperref}
\else
\usepackage[pagebackref=true,breaklinks=true,colorlinks,bookmarks=false]{hyperref}
\fi

\begin{document}

\title{Semi-Supervised Domain Adaptation with Auto-Encoder via Simultaneous Learning}

\author{Md Mahmudur Rahman\\
University of Massachusetts Lowell\\
{\tt\small mdmahmudur\_rahman@student.uml.edu}
\and
Rameswar Panda\\
MIT-IBM Watson AI Lab\\
{\tt\small rpanda@ibm.com}
\and
Mohammad Arif Ul Alam\\
University of Massachusetts Lowell\\
{\tt\small mohammadariful\_alam@uml.edu}
}

\maketitle
\thispagestyle{empty}

\begin{abstract}
   We present a new semi-supervised domain adaptation framework that combines a novel auto-encoder-based domain adaptation model with a simultaneous learning scheme providing stable improvements over state-of-the-art domain adaptation models. 
   Our framework holds strong distribution matching property by training both source and target auto-encoders using a novel simultaneous learning scheme on a single graph with an optimally modified MMD loss objective function.
   Additionally, we design a semi-supervised classification approach by transferring the aligned domain invariant feature spaces from source domain to the target domain. We evaluate on three datasets and show proof that our framework can effectively solve both fragile convergence (adversarial) and weak distribution matching problems between source and target feature space (discrepancy) with a high `speed' of adaptation requiring a very low number of iterations.
\end{abstract}

\section{Introduction}

\noindent Deep domain adaptation has become an emerging learning technique to address the lack of labeled data with the help of data from related but different domains\cite{tzeng2017adversarial, csurka2017domain,tzeng2017adversarial}. The key advantages of deep domain adaptation networks are the ability to train models from partially labeled or even unlabeled data with the help of related source domain by transferring knowledge among those domains.
Much progress has been done in deep domain adaptation focusing on Unsupervised Domain Adaptation (UDA) or Semi-supervised domain Adaptation(SSDA). Although UDA methods can be converted to SSDA methods by introducing supervision from labeled target samples, both methods cannot guarantee the inter-domain feature distribution matching, especially with the unlabeled samples \cite{kim2020attract} due to the presence of intra-domain feature space discrepancy resulting poor performance \cite{saito2019semi}.

\begin{figure}[t]
\begin{center}
    
  \includegraphics[width=0.8\linewidth]{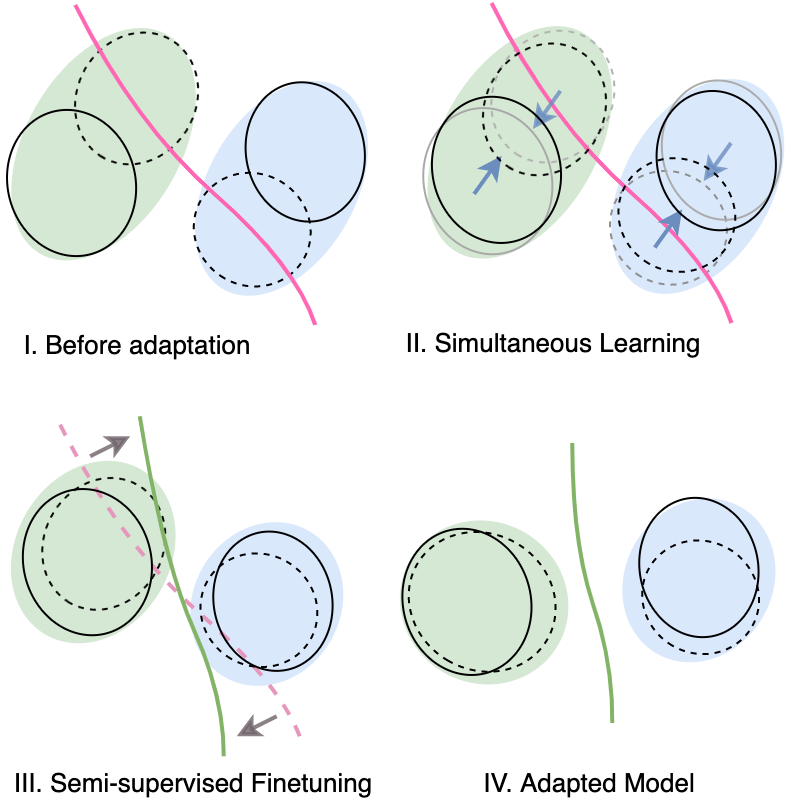} 
  \captionsetup{font=small}
  \vspace{-.1in}
  \caption{ Illustration of advantages of using our model. Here solid circle, dashed circle, solid line (red for before DA/green for after DA) represent source, target and decision boundary respectively. (a) decision boundary violates the feature space before adaptation. (b) While training with our framework iteratively, both of the feature spaces align each other. (c) Semi-supervised fine-tuning enables the decision boundaries to structure according to the newly adapted source and target feature spaces. Finally, (d) our method generates a robust decision boundary.}
  \label{fig:teaser} 
\end{center}
\end{figure}

\par
Based on the feature space mapping techniques, semi-supervised domain adaptation can be divided into three major categories: Discrepancy-based, adversarial discriminative, and adversarial generative models  \cite{csurka2017domain}. In adversarial discriminative models, a domain discriminator network tries to encourage domain confusion by optimizing an adversarial loss function \cite{tzeng2017adversarial}. A generative component is attached to the discriminative network to generate more similar feature spaces for source and target domains in the adversarial generative models \cite{goodfellow2014generative}. Although adversarial models achieve significant improvement, the training process of adversarial networks is complicated that often results very fragile convergence \cite{sam18}. Unlike the adversarial networks, a shared feature space has been learned by minimizing the distribution discrepancy between source and target domains in discrepancy-based methods \cite{long2017deep,zellinger2017central,sun2016return}. The main advantage of discrepancy-based methods is they are easy to train and convergence is not as fragile as adversarial networks \cite{CsurkaBCC17}. However, most discrepancy-based methods minimize the discrepancy between source and target domains and very few of them address discrepancy within domains \cite{abs-2008-06242,JiaWHZWH20,WangCWM20}. Especially when there is a low correlation between labeled and unlabeled samples in the target domain which results only partial alignment with the target feature space \cite{WangCWM20}. As a result, discrepancy-based methods can not guarantee distribution matching and often suffer from over-fitting to the target domain\cite{chen2020homm}.   

\par In this paper, we introduce a novel training concept called \emph{simultaneous learning} to address above shortcomings. Aligning to a fixed source feature space can be problematic because of the presence of an intra-domain discrepancy within the target domain \cite{kim2020attract}. To address this issue, we train both source and target network simultaneously where both feature spaces align one another in the course of learning (Figure \ref{fig:teaser}). We implement simultaneous learning with an auto-encoder-based domain adaptation where feature spaces of source and target domain get aligned in the bottleneck layer. In this framework, first, the simultaneous learning scheme allows the source domain to align with the target domain along with the intra-domain discrepancies present in the target domain. Concurrently, the target domain also gets aligned with the source domain with its intra-domain discrepancies. As a result, after learning iteratively, both source and target domain feature spaces align with each other including their intra-domain discrepancies, and solve the problem of partial alignment. Second, we use the bottleneck layer of auto-encoder as feature space rather than the layers in the classification network used in other methods. As the learning objective of the auto-encoder is not to classify samples, this enables a smooth transition and continuous feature space among the classes. If any feature discrepancy presents within the classes, continuous feature spaces help them to align completely with the other domain. 

To best of our knowledge, our proposed method is a first-of-its-kind framework that considers a simultaneous learning scheme with source and target networks with a carefully modified MMD loss across the domain invariant feature space. Our proposed framework is both easy to train and achieve state-of-the-art performance by solving the problem of partial alignment. {\bf key contributions}: 

\noindent $\bullet$ Propose a novel simultaneous learning scheme to help target feature space to align with the source domain feature space for SSDA.

\noindent $\bullet$ Modify MMD loss function in a way which works with simultaneous learning and achieve strong inter-domain alignment. 

\noindent $\bullet$ Propose an auto-encoder-based SSDA framework to address intra-domain discrepancy issue by implementing the simultaneous learning scheme and evaluate performances with state-of-the-art models using three datasets .

\section{Related Works}

{\it Unsupervised domain adaptation (UDA)} has become mostly popular recently that can be categorized in three approaches. First one is to reduce the distribution discrepancy by adversarial learning \cite{chen2020adversarial} \cite{wang2020classes} \cite{cui2020gradually} \cite{xu2020adversarial} which has few variances, such as utilization of adversarial learning to introduce an intermediate domain approach with Gradual Vanishing Bridge (GVB) layer between source and target domain \cite{cui2020gradually}. The second type of method minimizes the domain divergence with different loss functions, such as MMD \cite{long2017deep}, CORAL \cite{sun2016deep}, HoMM \cite{chen2020homm} etc. Recently, Domain Conditioned Adaptation Network (DCAN) \cite{li2020domain} uses domain conditioned channel attention mechanism to identify the domain specific features. The last type of method uses optimal transport (OT) mechanism which minimizes the cost of transporting one feature distribution to another domain \cite{chen2020graph} \cite{dhouib2020margin}\cite{xu2020reliable}\cite{li2020enhanced}. {\it Semi-supervised domain adaptation (SSDA)} fuses the well-labeled source domain with the label scarce target domain with the help of few labeled target samples. 
Kim et al. \cite{kim2020attract}  introduce the intra-domain discrepancy between the labeled and unlabeled data in the target domain and minimize it to align source and target domain features. Another approach uses meta-learning framework for multi-source and semi-supervised domain adaptation \cite{li2020online}. BiAT framework uses bidirectional adversarial training to guide the adversarial examples across the domain gap and achieve domain alignment \cite{jiangbidirectional}. Another adversarial learning based approach MME uses conditional minmax entropy optimization to achieve adaptation \cite{saito2019semi}.

\begin{figure*}[!htb]

\begin{center}
    \captionsetup{font=small}
  \includegraphics[width=\linewidth]{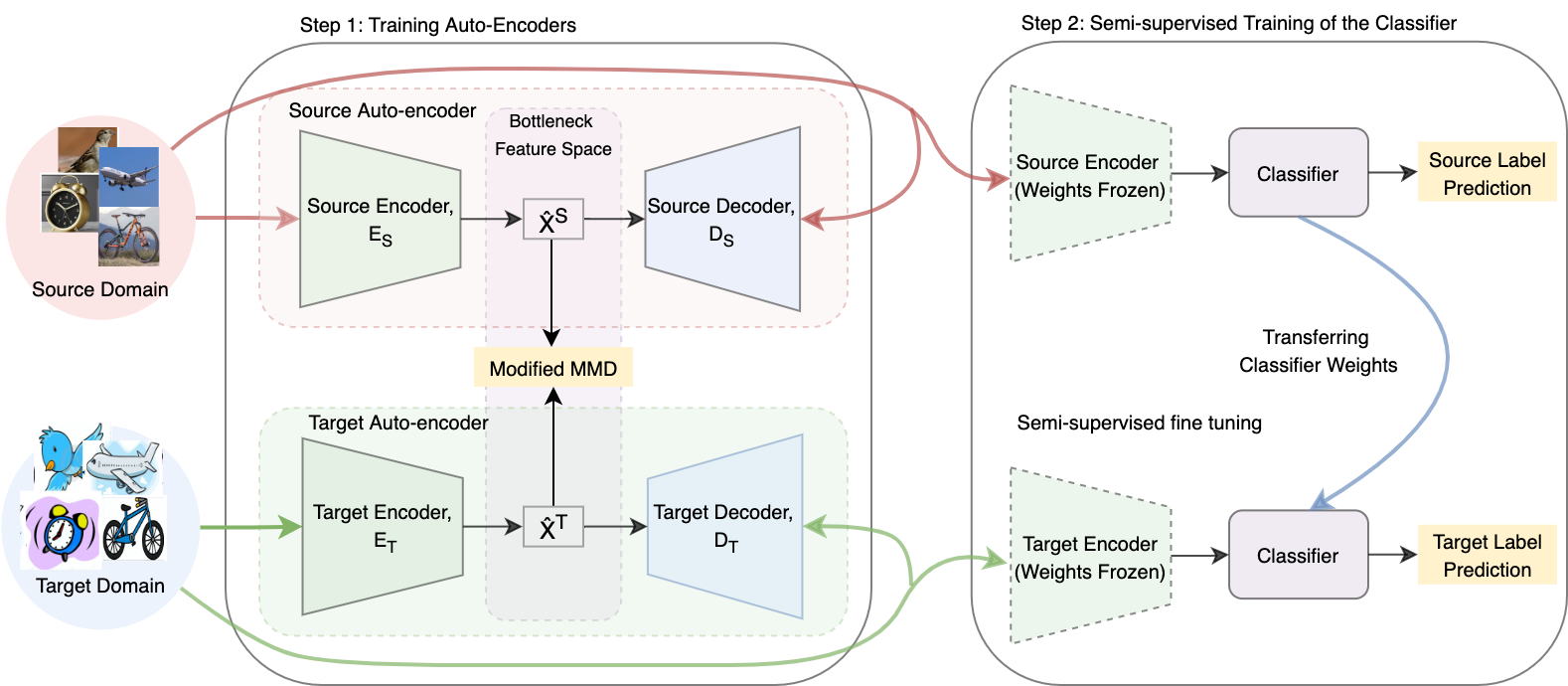}
  \vspace{-.1in}
  \caption{Training mechanism of our proposed Framework. Step 1: Training source and target auto-encoders with the source and target domain images respectively. $\Hat{X}^S$ and $\Hat{X}^S$ are the bottleneck feature space tied with modified MMD loss function. Step 2: Training the source encoder (weights are frozen) and classifier network with source images and corresponding labels. Then we attach the trained classifier with the target encoder (weight frozen) and fine tuning with labeled target images.}
  
  \label{fig:training_proc}

\end{center}
\end{figure*}

\section{Proposed Method}

Let the labeled source domain is $\mathcal{D}_s = \{(X_i^s, Y_i^s)\}^{n_s}$. Here $X_i^s = \{x_1, . . . , x_{n_s}\} \in \mathbb{R}^{d_s}$ is the $i^{th}$ sample from the source image with a feature space of $d_s$ dimension. Similarly, $Y_i^s \in \mathcal{Y}_s$ is the class label of corresponding $i^{th}$ sample of the source images $X_i^s$, where $\mathcal{Y}_s = \{1,2,...,n_c\}$ . 
Now, for our semi-supervised setup let consider the target domain $\mathcal{D}_t$ as union of two subset, labeled target domain, $\mathcal{D}_l$ and unlabeled target domain $\mathcal{D}_u$. So, $\mathcal{D}_t = \mathcal{D}_l \cup \mathcal{D}_u = \{(X_i^l, Y_i^l)\}^{n_l} \cup \{(X_i^u\}^{n_u}$ where $X_i^l \in \mathbb{R}^{d_t}$ is the labeled $i^{th}$ and $X_i^u \in \mathbb{R}^{d_t}$ is the unlabeled $i^{th}$ target image with feature space of $d_t$ dimension. $ Y_i^l \in \mathcal{Y}_t$ is the class label of corresponding $i^{th}$ labeled target data. In our semi-supervised domain adaptation setup we use three or one labeled samples per class for three-shot and one-shot experiments respectively, so, $n_u \gg n_l$ and $n_s \gg n_l$. The semi-supervised classification task in target domain, $\mathcal{T}_t$ can be defined to predict the class labels of the target domain, $Y^u_i$ with the unlabeled target data $X^u_i$. 
In our setting, we utilize both labeled ($\mathcal{D}_l$) and unlabeled ($\mathcal{D}_u$) target data to train the domain adaptation model for the task, $\mathcal{T}_t$. We assume that the source and target domain data consist of domain specific features $U^s$ , $U^t$ and domain invariant features $V^s$, $V^t$ respectively. The main goal is to map the domain invariant features of both domains into a common feature representation space which holds the common domain invariant features $V$ (where $V\approx V^s \approx V^t$) of both domains can represented as then, $P(\hat{X^s}|X^s) = P(\hat{X^t}|X^t) = V$ where $\hat{.}$ is feature representation space.

{\bf Overview}: Our system consists of three main components (figure \ref{fig:training_proc}): an auto-encoder for the source domain, an auto-encoder for the target domain, and the classifier. The training mechanism consists of two stages: training the auto-encoders, semi-supervised training of the classifier. At first, we train both source and target auto-encoders with their respective images. We implement our simultaneous learning scheme while training the auto-encoders to align the source and target bottleneck feature space one another with our modified MMD loss function along the course of learning. Second, we freeze the source encoder and attach a classifier layer after that. We train source encoder and classifier network with the source domain images with the supervision of corresponding labels. Finally, we attach the trained classifier to the target encoder and fine-tune with the labeled target domain samples which enables class alignment on already aligned feature space of source and target domain.

\subsection{Training Auto-encoders via Simultaneous Learning}
\begin{figure}
  \centering
  \includegraphics[width=\textwidth]{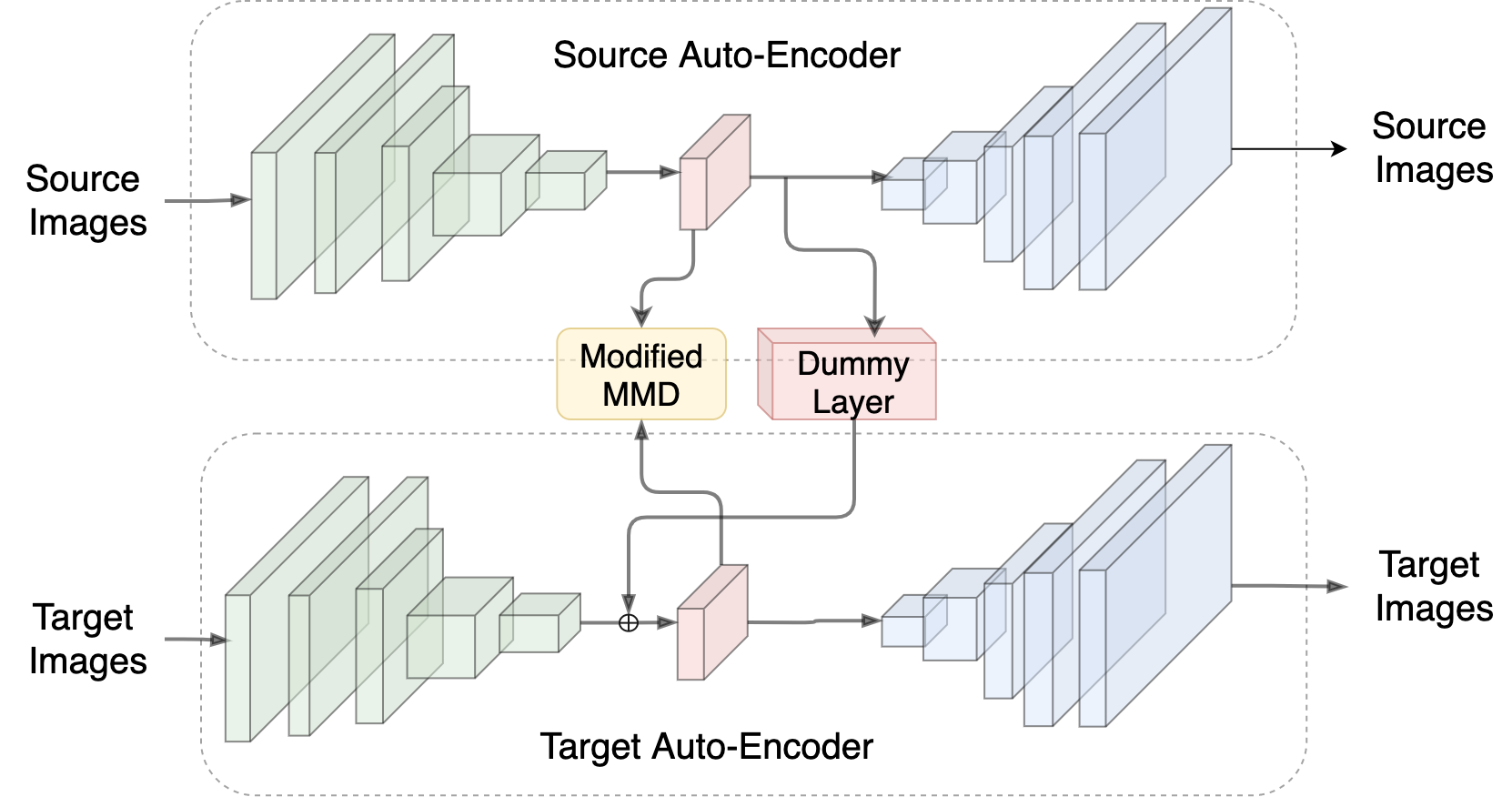}
  \begin{small}
  \caption{The network topology of the combined system of source and target auto-encoders is illustrated here where the optimization of modified MMD loss function minimizes the discrepancies between source and target bottleneck layers. The dummy layer connects both auto-encoders for simultaneous learning.}
  \label{fig:network_topology}
  \end{small}
\end{figure}

The objective of this stage is to train both of the auto-encoders with corresponding domain data.
During auto-encoder training, encoder part learns to create and embedding of input data and decoder part learns to reconstruct the same input sample from the embedding. At end, the source encoder, $E_s(\cdot)$ and the target encoder, $E_t(\cdot)$ learn to map the input image samples $X^s$ and $X^t$ to their corresponding feature spaces $\hat{X}^s$ and $\hat{X}^t$ respectively as $\hat{X}^s = E_s(X^s), \hat{X}^t = E_t(X^t)$. To facilitate our semi-supervised domain adaptation learning, we need a way to align the distribution of the embedding space of both encoders while training. So, the training objective is to minimize the distribution discrepancy between the bottleneck layers $\hat{X}^s$ and $\hat{X}^t$ along with reconstruction of the input image data. We achieve this objective we design a modified Maximum Mean Discrepancy (MMD) \cite{long2017deep} loss function to minimize the domain discrepancy in the bottleneck layers of the auto-encoder along the course of training. The source and target auto-encoders are fed with the source data $X^s$ and the target data $X^t$ respectively. We implement a simultaneous learning scheme by placing both auto-encoders in a single graph with a dummy layer. The dummy layer is a layer of nodes with zero activation. Therefore, it does not pass any information between the autoencoders but keeps both of them in a single graph. Figure \ref{fig:network_topology} shows the network topology of simultaneous learning scheme with the dummy layer. In this way, we can optimize both auto-encoders simultaneously in a single batch of data. The main advantage of simultaneous learning is the concurrent optimization of both source and target network. This concurrent optimization ensures matching of complex embedded features between two domains along the way of convergence. Thus, the self-reconstruction loss for source auto-encoder is:
\begin{small}
\begin{equation}
    \mathcal{L}_s = - \frac{1}{N_s} \sum_{i=1}^{N_s} X^s_i \cdot log(p(X^s_i)) + (1-X^s_i) \cdot log(1-p(X^s_i))
\vspace{-.05in}
\end{equation}
\end{small}
Here, $p(X^s_i)=E_s(D_s(X^s_i))$ is the probability of correct reconstruction of $X^s_i$ by the source auto-encoder and $N_s$ is the number of samples in the batch from source data.
On the other hand, the loss function of the target auto-encoder should consider both of the reconstruction loss and the optimization of bottleneck layer. We considered the weighted sum of self reconstruction loss and the discrepancy loss. In this case, we use modified MMD loss function as the discrepancy loss as follows:
\begin{small}
\begin{align}
    \mathcal{L}_t &= \mathcal{L}_{recon} + \beta \cdot \mathcal{L}_{md\_MMD} \\
    \mathcal{L}_{recon} &=  - \frac{1}{N_t} \sum_{i=1}^{N_t} X^t_i \cdot log(p(X^t_i)) + (1-X^t_i) \cdot log(1-p(X^s_i))
    \vspace{-.05in} 
    \end{align}
\end{small}

Here $\beta$ is the parameter for weighted sum in order to balance the comparative importance between the modified MMD loss and the reconstruction loss.

\noindent {\bf Modified MMD Loss Function}:
The Maximum Mean Discrepancy(MMD) \cite{gretton2007kernel} is a powerful metric for determining the statistical divergence between two correlated marginal probability distributions. The principal idea of MMD loss is to calculate the distance between the centroids of two different distributions. If we consider two bottleneck feature distributions of source and target auto-encoders, then we can define the MMD loss as
\begin{small}
\begin{equation}
\vspace{-.1in}
    \mathcal{L}_{MMD}(\hat{X}^s, \hat{X}^t) = \norm{\frac{1}{n_s}\sum_{i=1}^{n_s} \hat{X}^s_i - \frac{1}{n_t}\sum_{i=1}^{n_t} \hat{X}^t_i }^2
    \vspace{-.05in} 
    \end{equation}
\end{small}
Where $\hat{X}^s_i$ and $\hat{X}^t_i$ are the feature representations in the bottleneck layers for the $i^{th}$ sample of corresponding source and target domain data.
\par In designing our modified MMD loss, we intent to focus on the divergence between the class conditional probability distributions $\mathcal{P}(X|Y)$. The advantage of class conditional probability is that it focuses on optimizing the sample feature space based on classes rather the whole feature space at once (Equation \ref{eq:eq_md}). This is particularly helpful in the alignment where some discrepancies are present in the samples within the classes. As a result, the distance of class conditional probability can be estimated by calculating the centroid distance between corresponding source and target classes. So, our loss function becomes,
\begin{small}
\begin{equation}
        \label{eq:eq_md}
        \mathcal{L}_{md\_MMD}(\hat{X}^s, \hat{X}^t) = \frac{1}{C} \sum_{k=1}^C \norm{\frac{1}{n_s^k}\sum_{i=1}^{n_s^k} \hat{X}^s_{k,i} - \frac{1}{n_t^k}\sum_{i=1}^{n_t^k} \hat{X}^t_{k,i} }^2   
    \end{equation}
\end{small}
Here, the total number of classes is denoted by $C$, $n_s^k$ and $n_t^k$ is the number of source and target samples in $k$ class respectively. We use labelled source domain $\mathcal{D}_s = \{(X_i^s, Y_i^s)\}^{n_s}$ and labelled portion of target domain data $\mathcal{D}_l = \{(X_i^l, Y_i^l)\}^{n_l}$ to train with our modified MMD loss function in equation \ref{eq:eq_md}. Detailed training procedures are presented in the section 2 of supplementary material. 
\subsection{Semi-Supervised Training of the Classifier}

\noindent {\bf Simultaneous Learning}: We train both source and target autoencoders simultaneously with the parallel batches of source and target domain data. We made this possible by connecting both autoencoder networks to the same graph with a dummy layer. We build a data generator to generate the parallel data batches. The main advantage of simultaneous learning is, the source and target feature representation space can align with one another along the course of learning (Figure \ref{fig:teaser} II). Besides, with our modified MMD loss function, alignment occurs along similar classes of source and target domain. 
\par Most of the domain adaptation methods including CyCADA \cite{hoffman2018cycada} align along the intermediate layers of a classification stream network. On the other hand, our method aligns feature space in the bottleneck layer of an encoder decoder network. This helps to avoid class-wise clusterization in the feature representation space and transfer of features between the domains occurs all over the space rather than the clusters. Detailed discussion is presented in the section 1 of supplementary material.


The main function of the classifier network is to classify the bottleneck feature embedding. We use a common classifier network for both source and target feature embedding as we assume that both of the feature space would be aligned to each other before we apply classifier on them. Another important function of the common classifier is that it transfers the learning of classification from the source domain to the target domain. As the target domain have a small fraction of labeled data sample, the classifier network can learn robustly from the source domain where labeled samples are abound. Then, we fine-tune with the labeled target samples only for semi-supervised alignment. Considering the trained source and target auto-encoders, source and target encoders now learn to map the corresponding domain to a common feature space. We train the classifier network with the source feature space $\hat{X}^s$ and the labels of the source network with supervised training by cascading the learned source encoder and the classifier network. While training, we freeze the encoder module and let the classifier network be trained. So, the objective function while training the classifier with the source encoder is, $\min_{f_c} \mathcal{L}_c[Y^s, f_c(\hat{X}^s)]$ where, the classifier network is represented by $f_c(\cdot)$. We use categorical cross-entropy loss for learning the classifier network. Thus, loss function becomes, $\mathcal{L}_c = - \sum_{i=0}^C Y^s_i \cdot log(f_c(\hat{X}^s_i))$ The pseudo-code of the training procedures of our proposed framework is presented in Algorithm \ref{algo_1}. 

\begin{algorithm}[h]
\SetAlgoLined

\SetKwInOut{Input}{Input}\SetKwInOut{Output}{Output}
\SetKwRepeat{Do}{do}{while}
\Input{Source Domain Labeled data, $\mathcal{D}^s = \{X^s, Y^s\}$, Target Domain Labeled data, $\mathcal{D}^t_l = \{X^t_l, Y^t_l\}$, Unlabeled Target Domain, $\mathcal{D}^t_u =  \{X^t_u\}$, model parameter $\beta$, classifier layers size $c_l$, and bottleneck feature space size $b$}

\Output{End to end trained classifier prediction network for target domain unlabeled data}

Match the number of samples class by class between $\mathcal{D}^s$ and $\mathcal{D}^t_l$ by randomly resampling the smaller domain\;
Sort $\mathcal{D}^s$ and $\mathcal{D}^t_l$ by class.
Initialize the layer weights of the source and the target auto-encoder randomly\;
Set the corresponding loss functions $\mathcal{L}_s$ and $\mathcal{L}_t$ to the source and the target auto-encoder respectively\;
\Repeat{$\mathcal{L}_s$ and $\mathcal{L}_t$ converges}{
Update weights of the source and target auto-encoders with batches of data from $\{X^s, X^s\}$ and $\{X^t_l, X^t_l\}$ respectively. 
}

Take only Encoder module of source Auto-encoder network and cascade the Classifier network at the end of it\;
Freeze the Encoder weights and randomly initialize the weights of the Classifier network\;
\Repeat{Validation Loss converge}{
Optimize the "Source Encoder + Classifier" network with $X^s, Y^s$\;
}

Take The Classifier module and cascade with the target encoder\;
Freeze the target encoder weights\;
\Repeat{Validation loss converge}{
Optimize the "Target Encoder + Classifier" network with labelled target data, $\{X^t_l, Y^t_l\}$\;
}
Estimate the label of the unlabeled samples from target domain, $\mathcal{D}^t_u =  \{X^t_u\}$ with "Target Encoder + Classifier" network\;

 \caption{Training Procedures of Auto-encoder based Semi-Supervised Domain Adaptation}
 \label{algo_1}
\end{algorithm}

\section{Experiments}
 

\begin{table}[t]
    \small
    \captionsetup{font=small}
    \caption{Results (\% Accuracy on target) on Office31 data with 3-shot setting on 3 domains, D: DSLR, A: Amazon, W: Webcam }
    \centering
    \label{result-table-office}
    \begin{tabular}{lll|ll|l}
    \toprule
    Backbone & \multicolumn{2}{c|}{Alexnet} & \multicolumn{2}{c|}{VGG16}\\
    \midrule
    Method     & D$\rightarrow$A   & W$\rightarrow$A & D$\rightarrow$A   & W$\rightarrow$A & Avg\\
    \midrule
    S+T & $62.4$ & $61.2$ & $73.3$ & $73.2$ & 67.5\\
    DANN & $65.2$  & $64.4$ & $74.6$ & $75.4$ & 69.9\\
    ADR & $61.4$  & $61.2$ & $74.1$ & $73.3$ & 67.5\\
    CDAN & $61.4$ & $60.3$ & $71.4$ & $74.4$ & 66.9\\
    MME & $67.8$ & $67.3$ & $\mathbf{77.6}$ & $76.3$ & 72.3\\
    APE & $69.0$ & $67.6$ & $76.6$ & $75.8$ & 72.3\\
    \textbf{Our Method} & $\mathbf{70.5}$ & $\mathbf{69.7}$ & $77.4$ & $\mathbf{77.8}$ & $\mathbf{73.8}$\\
    
    \bottomrule
    \end{tabular}
\end{table}

We use three widely used benchmark datasets: \textbf{DomainNet}\cite{peng2019moment} is a recent large scale domain adaptation dataset consisting 6 different domains and 345 classes. As the labels of some domains and classes are very noisy \cite{saito2019semi}, we use a subset of 126 most frequent classes and four domains (Real, Painting, Clip-art and Sketch) for our experiments. This subset has around 36500 images per domain. We use a total of 7 different domain adaptation scenarios for evaluating our model. For the choice of 4 domains and 7 domain adaptation use cases for DomainNet dataset, we follow APE\cite{kim2020attract}, BiAT\cite{jiangbidirectional} and MME\cite{saito2019semi} for easier comparison. \textbf{Office-31} is a commonly used domain adaptation dataset comprising 4110 images of 31 classes from everyday office environment in three distinct domains: \textbf{A}mazon (images from amazon.com), \textbf{D}SLR (High resolution image taken with DSLR camera) and \textbf{W}ebcam (Low-resolution images taken with webcam). We evaluate our method with all six transfer task combining three of the domains. \textbf{Digit Recognition Datasets} consists of five widely used digit datasets: \textbf{MNIST} \cite{lecun1998gradient}, \textbf{SVHN} \cite{netzer2011reading}, \textbf{USPS} \cite{hull1994database}, \textbf{Synthetic digit dataset} \cite{ganin2016domain}. All of the datasets have 10 classes of 10 digits. We have evaluated the performance over the following pairs of cross-domain adaptation: USPS $\xrightarrow{}$ MNIST, SVHN $\xrightarrow{}$ MNIST and Synthetic Digits $\xrightarrow{}$ MNIST. All of the datasets have nearly similar number of samples and numbers of pixels of every sample is also similar.

\begin{table*}[t]

\small
\captionsetup{font=small}
  \caption{Results (\% Accuracy on target) on DomainNet dataset with three-shot setting (three samples per class in the target domain) on 4 domains, R: Real, C: Clipart, P: Painting, S: Sketch. Our proposed method outperforms other state of the art methods on most domain adaptation scenarios}
 
  \label{result-table-domainnet}
  \centering
  \begin{tabular}{p{0.5cm}p{1.6cm}p{0.4cm}p{0.4cm}p{0.4cm}p{0.4cm}p{0.4cm}p{0.4cm}p{0.4cm}p{0.4cm}|p{0.4cm}p{0.4cm}p{0.4cm}p{0.4cm}p{0.4cm}p{0.4cm}p{0.4cm}p{0.4cm}}
    \toprule
    - & Backbone & \multicolumn{8}{c|}{Alexnet} & \multicolumn{8}{c}{ResNet34}   \\
    \midrule
    Shot & Method     & R$\rightarrow$C  & R$\rightarrow$P  & P$\rightarrow$C & C$\rightarrow$S & S$\rightarrow$P & R$\rightarrow$S & P$\rightarrow$R & Avg & R$\rightarrow$C  & R$\rightarrow$P  & P$\rightarrow$C & C$\rightarrow$S & S$\rightarrow$P & R$\rightarrow$S & P$\rightarrow$R & Avg\\
    \midrule
    \multirow{8}{*}{1-shot}& S+T & $43.3$ & $42.4$ & $40.1$ & $33.6$ & $35.7$ & $29.1$ & $55.8$ & $40.0$ & $55.6$ & $60.6$ & $56.8$ & $50.8$ & $56.0$ & $46.3$ & $71.8$ & $56.8$\\
    &DANN\cite{ganin2016domain} & $43.3$ & $41.6$ & $39.1$ & $35.9$ & $36.9$ & $32.5$ & $53.6$ & $40.4$ & $58.2$ & $61.4$ & $56.3$ & $52.8$ & $57.4$ & $52.2$ & $70.3$ & $58.4$\\
    &ADR\cite{saito2017adversarial} & $43.1$  & $41.4$ & $39.3$ & $32.8$  & $33.1$ & $29.1$ & $55.9$ & $39.2$ & $57.1$ & $61.3$ & $57.0$ & $51.0$ & $56.0$ & $49.0$ & $72.0$ & $57.6$\\
    &CDAN\cite{long2017conditional} & $46.3$ & $45.7$ & $38.3$ & $27.5$ & $30.2$ & $28.8$ & $56.7$ & $39.1$ & $65.0$ & $64.9$ & $63.7$ & $53.1$ & $63.4$ & $54.5$ & $73.2$ & $62.5$\\
    &MME\cite{saito2019semi}  & $48.9$ & $48.0$ & $46.7$ & $36.3$ & $39.4$ & $33.3$ & $56.8$ & $44.2$ & $70.0$ & $67.7$ & $69.0$ & $56.3$ & $64.8$ & $61.0$ & $76.1$ & $66.4$\\
    &APE\cite{kim2020attract}  & $47.7$ & $49.0$ & $46.9$ & $38.5$ & $38.5$ & $33.8$ & $\mathbf{57.5}$ & $44.6$ & $70.4$ & $70.8$ & $72.9$ & $56.7$ & $64.5$ & $63.0$ & $76.6$ & $67.8$\\
    &BiAT\cite{jiangbidirectional} & $54.2$ & $49.2$ & $44.0$ & $37.7$ & $39.6$ & $37.2$ & $56.9$ & $45.5$ & $73.0$ & $68.0$ & $71.6$ & $57.9$ & $63.9$ & $58.5$ & $77.0$ & $67.1$\\
    &\textbf{Our Method} & $\mathbf{56.3}$ & $\mathbf{50.5}$ & $\mathbf{48.8}$ & $\mathbf{40.2}$ & $\mathbf{41.5}$ & $\mathbf{39.3}$  & $58.4$ & $\mathbf{47.8}$ & $\mathbf{74.8}$ & $\mathbf{72.4}$ & $\mathbf{74.6}$ & $\mathbf{59.5}$ & $\mathbf{66.6}$ & $\mathbf{66.2}$ & $\mathbf{79.4}$ & $\mathbf{70.5}$\\
    \midrule
    \multirow{8}{*}{3-shot}& S+T & $47.1$ & $45.0$ & $44.9$ & $36.4$ & $38.4$ & $33.3$ & $58.7$ & $43.4$ & $60.0$ & $62.2$ & $59.4$ & $55.0$ & $59.5$ & $50.1$ & $73.9$ & $60.0$\\
    &DANN & $46.1$ & $43.8$ & $41.0$ & $36.5$ & $38.9$ & $33.4$ & $57.3$ & $42.4$ & $59.8$ & $62.8$ & $59.6$ & $55.4$ & $59.9$ & $54.9$ & $72.2$ & $60.7$\\
    &ADR & $46.2$  & $44.4$ & $43.6$ & $36.4$  & $38.9$ & $32.4$ & $57.3$ & $42.7$ & $60.7$ & $61.9$ & $60.7$ & $54.4$ & $59.9$ & $51.1$ & $74.2$ & $60.4$\\
    &CDAN & $46.8$ & $45.0$ & $42.3$ & $29.5$ & $33.7$ & $31.3$ & $58.7$ & $41.0$ & $69.0$ & $67.3$ & $68.4$ & $57.8$ & $65.3$ & $59.0$ & $78.5$ & $66.5$\\
    &MME  & $55.6$ & $49.0$ & $51.7$ & $39.4$ & $43.0$ & $37.9$ & $60.7$ & $48.2$ & $72.2$ & $69.7$ & $71.7$ & $61.8$ & $66.8$ & $61.9$ & $78.5$ & $68.9$\\
    &APE  & $54.6$ & $50.5$ & $52.1$ & $42.6$ & $42.2$ & $38.7$ & $\mathbf{61.4}$ & $48.9$ & $76.6$ & $72.1$ & $\mathbf{76.7}$ & $63.1$ & $66.1$ & $67.8$ & $79.4$ & $71.7$\\
    &BiAT & $58.6$ & $50.6$ & $52.0$ & $41.9$ & $42.1$ & $42.0$ & $58.8$ & $49.4$ & $74.9$ & $68.8$ & $74.6$ & $61.5$ & $67.5$ & $62.1$ & $78.6$ & $69.7$\\
    &\textbf{Our Method} & $\mathbf{59.1}$ & $\mathbf{52.7}$ & $\mathbf{54.6}$ & $\mathbf{44.5}$ & $\mathbf{43.9}$ & $\mathbf{45.3}$  & $60.1$ & $\mathbf{51.5}$ & $\mathbf{77.3}$ & $\mathbf{74.1}$ & $76.2$ & $\mathbf{65.2}$ & $\mathbf{69.6}$ & $\mathbf{69.5}$ & $\mathbf{80.5}$ & $\mathbf{73.2}$\\
    \bottomrule
    
  \end{tabular}
\end{table*}

\begin{table}[t]
    \small
    \centering
    \vspace{-.2in} 
    \captionsetup{font=small}
    \caption{Results (\% Accuracy on target) on digit recognition dataset with three-shot learning setting, S: SVHN, M: MNIST, Syn: Synthetic Digit dataset}
  \label{result-table-digit}
  \begin{tabular}{lllll}
    \toprule
    Method     & S$\rightarrow$M   & U$\rightarrow$M & Syn$\rightarrow$M & Avg\\
    \midrule
    Modified LeNet & $67.3$ & $66.4$ & $89.7$ & 74.5\\
    DDC \cite{tzeng2014deep} & $71.9$ & $75.8$ & $89.9$  & 79.2\\
    DAN \cite{long2015learning} & $79.5$  & $89.8$ & $75.2$ & 81.5\\
    DANN \cite{ganin2016domain} & $70.6$  & $76.6$ & $90.2$ & 79.1\\
    CMD \cite{zellinger2017central} & $86.5$  & $86.3$ & $96.1$ & 89.6\\
    ADDA \cite{tzeng2017adversarial} & $72.3$  & $92.1$ & $96.3$ & 86.9\\
    CORAL \cite{sun2016deep} & $89.5$  & $\mathbf{96.5}$ & $96.5$ & 91.0\\
    \textbf{Our Method} & $\mathbf{95.22}$ & $93.8$ & $\mathbf{96.6}$ & $\mathbf{92.8}$\\
    
    \bottomrule
  \end{tabular}
    
\end{table}

\subsection{Implementation Details}

We implemented our proposed model with python Keras (tensorflow backend). We resize all of the input images into $224\times224$ pixels for the three channel images of office-31 and DomainNet datasets and $32\times32$ pixels images for the digit recognition datasets. We implement Alexnet and ResNet34 backbone for the DomainNet dataset; Alexnet and VGG16 for the Office-31 dataset and VGG16 network for the digit recognition dataset in our encoder module. In case of decoder, we use the same encoder layers in reversed order. We use the Upsampling layer of same size in the decoder in place of the max-pool layer in the encoder to maintain the progressively increasing shape of decoder. We also add a reconstruction layer of the shape of input image at the end of the decoder. We optimize the system using Adam \cite{kingma2014adam} optimization function with learning rate of $1\times10^{-3}$. We choose the optimized values the learning rate and weighting the parameter $\beta$ (=0.25) using hyper-parameter tuning (see implementation details in supplementary material).

\par We run the learning model of our framework on a server having a cluster of three Nvidia GTX GeForce Titan X GPU and Intel Xeon CPU (2.00GHz) processor with 12 Gigabytes of RAM.

\begin{figure}
  \centering
  \captionsetup{font=small}
  \includegraphics[width=\linewidth]{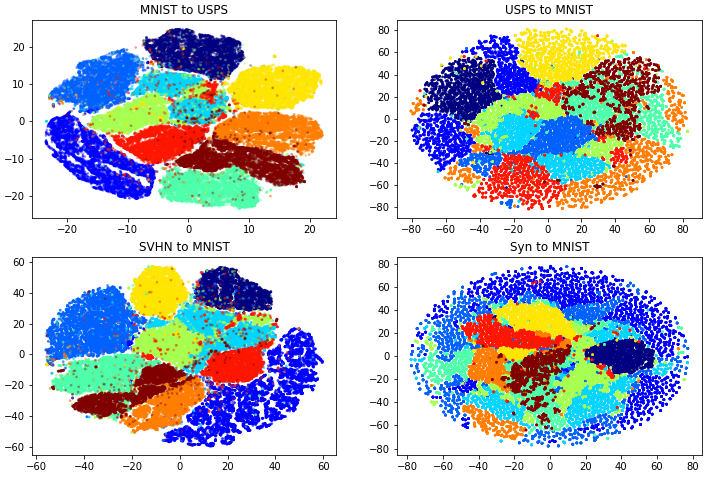}
  
  \caption{
  t-SNE Visualization of the bottleneck layer feature representation of the target domain. 
  }
  \label{rep_s}
\end{figure}

    



\begin{figure}
\begin{center}
\captionsetup{font=small}
  \includegraphics[width=\linewidth]{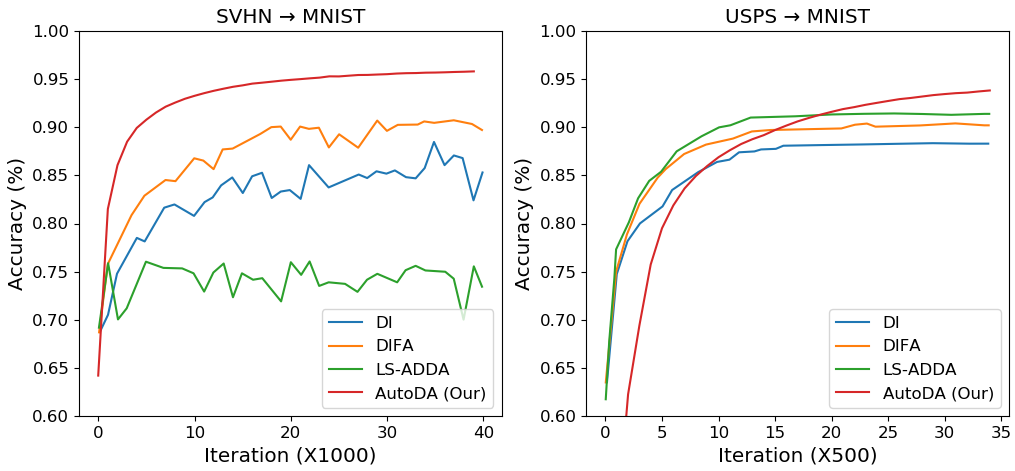}
  \caption{Comparison of Our method's convergence performance with baseline algorithms LS-ADDA (Least Square ADDA), DI (Domain Invariance LS-ADDA) and DIFA(Domain Invariance and Feature Augmentation LS-ADDA)}
  \label{convergence}
\end{center}
\end{figure}

We compare the performance of our framework with recent SSDA frameworks such as APE\cite{kim2020attract}, BiAT\cite{jiangbidirectional} and MME\cite{saito2019semi}, DANN\cite{ganin2016domain}, CDAN\cite{long2017conditional} and ADR \cite{saito2017adversarial,saito2019semi}. We considered the baseline method as "S+T" where all of the available labeled source and target data are used to train the network. Table \ref{result-table-domainnet} shows details of performance comparisons on {\bf DomainNet dataset} where our framework outperforms nearly all of the current state-of-the-art methods by 4-8\% margin with different backbones both in 1-shot and 3-shot scenario. On average our method outperforms baseline methods by 7.8\% in case of 1-shot and 13.7\% in case 3-shot learning with Alexnet backbone, 8.1\% in 1-shot and 13.2\% in 3-shot learning with ResNet34 backbone network. Some of the state of the art method, DANN, ADR and CDAN performs poorly compared to the baseline accuracy score specially in the 1-shot learning. In some cases they even suffer from negative learning while domain transfer. Our method shows significant performance improvement. However, the performance improvement is larger in 3-shot scenario than 1-shot. Table \ref{result-table-office} shows the performance comparison of our model on {\bf Office 31 dataset} where we can see our model outperforms all of the existing methods with a significant margin with both Alexnet and VGG16 backbone in almost all domain adaptation scenarios. Here, ADR and CDAN shows similar or degenerative performance on the target dataset compared to the baseline because of potential negative transfer. On the other hand, our method outperformed the baseline method by 5-10\% margin. Finally, the performance comparison of our proposed algorithm on the {\bf digit recognition datasets} with some state-of-art adversarial network- and discrepancy-based methods is presented on the table \ref{result-table-digit}. We compare the accuracy score on target using the performance of modified LeNet network as baseline. Our framework outperformed all other state-of-the-art methods in all of the domain adaptation scenarios except CORAL in USPS$\rightarrow$MNIST. It is worth to note that, in case of SVHN$\rightarrow$MNIST, our model outperformed all other algorithms with a significant margin where source and target images are mostly dissimilar.

\par The performance of the discrepancy based methods (DDC, DAN, CMD, CORAL) and adversarial domain adaptation methods (DANN, ADDA) are significantly superior than the baseline Convolution neural network based Modified LeNet model. But their performance varies widely among different domain adaptation scenarios which shows their lack of robustness with the variation of domains. However, our framework outperforms all of the baseline methods on nearly all of the domain adaptation scenarios as well as on the average performance. This confirms the ability of our framework to minimize the domain discrepancy efficiently with a simple and stable training procedure.

\begin{figure}
\begin{center}
\captionsetup{font=small}
    
  \includegraphics[width=\linewidth]{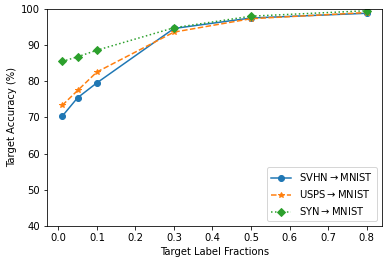}
  \caption{Domain adaptation performance changes on different sample size (fraction of samples) of target data}
  \label{fig:fraction_convergence}
\end{center}
\end{figure}

\subsection{Analysis}

\noindent {\bf Ablation Studies}: We perform ablation studies on DomainNet and Office31 dataset with Resnet34 backbone to understand performance changes over different loss functions and hyperparameters.

    
   


\noindent {\it Effect of proposed Modified MMD loss function and simultaneous learning}: We perform an experiment by removing the Modified MMD loss function to observe its effect on domain alignment. The drop of target accuracy score shows the importance of our modified MMD loss function (Table \ref{tab:ablation}). We also use sequential learning instead of simultaneous learning and observe a drop of target accuracy score in both datasets.

\begin{figure}[]
\captionsetup{font=small}
    \centering
    \ffigbox{%
    \includegraphics[width=\linewidth]{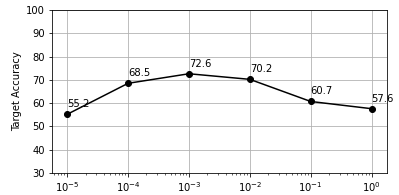}
    }{%
    \caption{Effect of $\beta$(Modified MMD loss function) on DomainNet data}
    \label{fig:ablbeta}
    }
\end{figure}

\noindent  {\it Effect of Hyperparameters}: We analyze the effect of different weights of modified MMD loss function ($\beta$) over auto-encoder reconstruction loss with DomainNet dataset (figure \ref{fig:ablbeta}). Target accuracy score is plotted against the logarithmic value of $\beta$. We observe that the $\beta$ value of $10^{-3}$ results maximum accuracy in target domain. 

\begin{table}[H]
\captionsetup{font=small}
\small
    \centering
      \begin{tabular}{c|c|c}
        \hline
        Approach &  DomainNet &  Office 31 \\
        \hline
        w/o Modified MMD Loss & $65.4\pm0.34$ & $67.5\pm0.4$ \\
        w/o Simultaneous Learning & $61.7\pm0.56$ & $62.4\pm0.35$ \\
        \hline
        Our Method & $77.3\pm0.25$& $73.7\pm0.3$ \\
        \hline
    \end{tabular}
  \caption{Effect of modified MMD loss function and simultaneous learning on DomainNet data}%
  \label{tab:ablation}
\end{table}

\noindent {\bf Bottleneck Space visualization}: We visualize the bottleneck representation space of the auto-encoders of our proposed framework after the completion of training the auto-encoders. We randomly select 2000 samples from MNIST and USPS data-set and get 100 dimensional representation space from the bottleneck layer. We use t-SNE \cite{van2008visualizing} visualization method to extract two principal components from that 100 dimensional space keeping the t-SNE parameters consistent.
\par In figure \ref{rep_s}, we observe a smooth transition between different classes rather than clustered centroids yet with a clear distinction which helps to obtain domain adaptation along a wide region of bottleneck representation space.

\noindent {\bf  Convergence Performance}: To compare the stability of our system, we compared the convergence performance of our model with three variations of ADDA \cite{tzeng2017adversarial} described in \cite{Volpi_2018_CVPR}: LS-ADDA (Least Square ADDA), DI (Domain Invariance LS-ADDA) and DIFA(Domain Invariance and Feature Augmentation LS-ADDA), as shown in Fig. \ref{convergence}. In case of both SVHN$\rightarrow$MNIST and USPS$\rightarrow$MNIST, our model showed higher stability and higher score. In case of SVHN$\rightarrow$MNIST, our model converged way faster than all three other variations of ADDA. This validates the theoretical assumption of higher stability of our model compared to adversarial network-based methods. Fig. \ref{fig:fraction_convergence} shows the convergence of our model with the variations of the target labels (fraction) which provides ample proof that our method adapts fast enough to be considered as a simple solution against the complexity of training and uncertainty of domain invariant feature spaces of other methods.

\section{Limitations}
However, if there is high class imbalance present between source and target domain, our proposed method as well as other discrepancy based methods may perform poorly because the discrepancy property does not hold strongly in case of imbalance data. However, we have tried to overcome this issue by randomly sampling the minor classes while using the class-wise MMD loss function. But this can lead to over fitting in highly imbalanced data. Besides, the auto-encoder bottleneck layer need to be designed optimally. Too narrow bottleneck layer can wash away some important information necessary to classification which can lead to lower classification accuracy. On the other hand, too wide bottleneck layer can make slow convergence of the modified MMD loss function which can lead to slower or failed domain adaptation. In future, we will expand our domain adaptation framework to work with multiple domain adaptation problem which is under-explored due to training difficulties. We also have a plan to implement adversarial based techniques to minimize domain discrepancy between bottleneck layers besides Class-wise MMD loss function.

\section{Conclusion}
We propose an auto-encoder based semi-supervised domain adaptation framework and introduce a novel simultaneous learning scheme for source and target domain. We also introduce a modified version of Maximum Mean Discrepancy (MMD) loss function and implement it to minimize domain discrepancy. Our experimental evaluation and further analysis show the outperformance of our method over state-of-art and its theoretical validity. To induce further superiority of our proposed model, we present evaluation across four domain shifts which provides ample proof that our simultaneous learning scheme would make it easier to train multiple domains at a time.

\section{Acknowledgment}
This work is partially supported by NSF’s Smart \& Connected Community award \#2230180

{\small
\bibliographystyle{ieee_fullname}
\bibliography{egbib}
}

\end{document}